\documentclass[runningheads]{llncs}

\usepackage[title]{appendix}

\makeatletter%
\newcommand{\@chapapp}{\relax}%
\makeatother%

\usepackage[T1]{fontenc}
\usepackage{amsmath}
\usepackage{booktabs}
\usepackage{multirow}
\usepackage{tikz}
\usetikzlibrary{positioning}
\usepackage{graphicx}
\usepackage{hyperref}
\usepackage{microtype}

\begin{document}
\title{Learning Generative Models for Active Inference using Tensor Networks}
\author{Samuel T.\ Wauthier\inst{1} \and
Bram Vanhecke\inst{2} \and
Tim Verbelen\inst{1} \and
Bart Dhoedt\inst{1}}
\authorrunning{S.\ T.\ Wauthier et al.}
\institute{
IDLab, Department of Information Technology at Ghent University -- imec,
Technologiepark-Zwijnaarde 126, B-9052 Ghent, Belgium \\
\email{\{firstname.lastname\}@ugent.be}
\and
University of Vienna, Faculty of Physics and Faculty of Mathematics, Quantum~Optics, Quantum~Nanophysics~and~Quantum~Information,\\ Boltzmanngasse 5, 1090 Vienna, Austria \\
\email{bram.andre.roland.vanhecke@univie.ac.at}
}
\maketitle
\begin{abstract}
Active inference provides a general framework for behavior and learning in autonomous agents.
It states that an agent will attempt to minimize its variational free energy, defined in terms of beliefs over observations, internal states and policies.
Traditionally, every aspect of a discrete active inference model must be specified by hand, i.e.\ by manually defining the hidden state space structure, as well as the required distributions such as likelihood and transition probabilities.
Recently, efforts have been made to learn state space representations automatically from observations using deep neural networks.
% However, these models are typically overparameterized, with the risk of overfitting the data at hand.
In this paper, we present a novel approach of learning state spaces using quantum physics-inspired tensor networks.
The ability of tensor networks to represent the probabilistic nature of quantum states as well as to reduce large state spaces makes tensor networks a natural candidate for active inference.
We show how tensor networks can be used as a generative model for sequential data.
Furthermore, we show how one can obtain beliefs from such a generative model and how an active inference agent can use these to compute the expected free energy.
Finally, we demonstrate our method on the classic T-maze environment.
\end{abstract}

\section{Introduction}
% active inference needs to be manually defined
Active inference is a theory of behavior and learning in autonomous agents~\cite{Friston2016}.
An active inference agent selects actions based on beliefs about the environment in an attempt to minimize its variational free energy.
As a result, the agent will try to reach its preferred state and minimize its uncertainty about the environment at the same time.

This scheme assumes that the agent possesses an internal model of the world and that it updates this model when new information, in the form of observations, becomes available.
In current implementations, certain components of the model must be specified by hand.
For example, the hidden space structure, as well as transition dynamics and likelihood, must be manually defined.
Deep active inference models deal with this problem by learning these parts of the model through neural networks~\cite{Ueltzhoffer2018,Catal2020}.
% Unfortunately, the models are typically overparameterized at the risk of overfitting on the training data.
% Unfortunately, theoretical understanding of deep neural network models is lacking and they are often seen as a ``black box'' approach.

Tensor networks, as opposed to neural networks, are networks constructed out of contractions between tensors.
In recent years, tensor networks have found their place within the field of artificial intelligence.
More specifically, Stoudenmire and Schwab~\cite{Stoudenmire2016} showed that it is possible to train these networks in an unsupervised manner to classify images from the MNIST handwritten digits dataset~\cite{mnist}.
Importantly, tensor networks have shown to be valuable tools for generative modeling.
Han et al.~\cite{Han2018} and Cheng et al.~\cite{Cheng2019} used tensor networks for generative modeling of the MNIST dataset, while Mencia~Uranga and Lamacraft \cite{MenciaUranga2020} used a tensor network to model raw audio.

% benefits of tensor networks
Tensor networks, which were originally developed to represent quantum states in many-body quantum physics, are a natural candidate for generative models for two reasons.
Firstly, they were developed in order to deal with the curse of dimensionality in quantum systems, where the dimensionality of the Hilbert space grows exponentially with the number of particles.
Secondly, they are used to represent quantum states and are, therefore, inherently capable of representing uncertainty, or, in the case of active inference, beliefs.
For example, contrary to neural networks, tensor networks do not require specifying a probability distribution for the hidden state variables or output variables.
% Furthermore, tensor networks provide a potent tool to tackle theoretical understanding of
% approaches are advantageous in terms of theoretical understanding, as their expressive power is relatively well-understood using entanglement properties~\cite{Liu2021}.
Furthermore, tensor networks can be exactly mapped to quantum circuits, which is important for the future of quantum machine learning~\cite{tns}.

In this paper, we present a novel approach to learning state spaces, likelihood and transition dynamics using the aforementioned tensor networks.
We show that tensor networks are able to represent generative models of sequential data and how beliefs (i.e.\ probabilities) about observations naturally roll out of the model.
Furthermore, we show how to compute the expected free energy for such a model using the sophisticated active inference scheme.
We demonstrate this using the classic T-maze environment.

Section~\ref{ch:tn} elaborates on the inner workings of a tensor network and explains how to train such a model.
Section~\ref{ch:environment} explains the environment and how we applied a tensor network for planning with active inference.
In section~\ref{ch:results}, we present and discuss the results of our experiments on the T-maze environment.
Finally, in section \ref{ch:conclusion}, we summarize our findings and examine future prospects.

\section{Generative modeling with tensor networks}\label{ch:tn}
% As the name suggests, tensor networks consist of networks of tensors.
% Their building blocks, tensors, can be seen as the objects that generalize objects such as vectors and matrices.
% Networks are constructed by linking tensors through contractions.
% Contractions, in turn, can be seen as the operations that generalize operations such as matrix products, inner products, and traces.
% More mathematically, a contraction is the summation over a shared index, e.g.\ matrix multiplication
% $\sum_j A_{ij} B_{jk}$, inner product $\sum_i v_i w_i$, and trace of a matrix $\sum_i A_{ii}$.

% Graphically, a tensor can be represented as node with a number of free edges equal to its number of indices, e.g.\ a matrix is a node with 2 free edges, while a vector has a single free edge. [insert figure]
% A contraction is represented by connecting free nodes.
% Connecting nodes in this way, it is possible to construct networks of connected tensors, which symbolize the summation over shared indices between a set of tensors.

A generative model is a statistical model of the joint probability $P(X)$ of a set of variables $X = (X_1, X_2, \ldots, X_n)$.
% Learning a generative model requires a data set which is representative of the underlying probability distribution.
As previously mentioned, quantum states inherently contain uncertainty, i.e.\ they embody the probability distribution of a measurement of a system.
It is natural, then, to represent a generative model as a quantum state.
Quantum states can be mathematically described through a wave function $\Psi(x)$ with $x = (x_1, x_2, \ldots, x_n)$ a set of real variables, such that the probability of $x$ is given by the Born rule:
\begin{equation}
    P(X = x) = \frac{|\Psi(x)|^2}{Z},
\end{equation}
with $Z = \sum_{\{x\}} |\Psi(x)|^2$, where the summation runs over all possible realizations of the values of $x$.

Recent work~\cite{Orus2019} has pointed out that quantum states can be efficiently parameterized using tensor networks.
The simplest form of tensor network is the matrix product state (MPS), also known as a tensor train~\cite{Perez-Garcia2007}.
When representing a quantum state as an MPS, the wave function can be parameterized as follows:
\begin{equation}\label{eq:mps}
    \Psi(x) = \sum_{\alpha_1} \sum_{\alpha_2} \sum_{\alpha_3} \ldots \sum_{\alpha_{n-1}} A^{(1)}_{\alpha_1}(x_1) A^{(2)}_{\alpha_1\alpha_2}(x_2) A^{(3)}_{\alpha_2\alpha_3}(x_3) \cdots A^{(n)}_{\alpha_{n-1}}(x_n).
\end{equation}
Here, each $A^{(i)}_{\alpha_{i-1}\alpha_i}(x_i)$ denotes a tensor of rank 2 (aside from the tensors on the extremities which are rank 1) which depends on the input variable $x_i$.
This way, the wave function $\Psi(x)$ is decomposed into a series of tensors $A^{(i)}$.

Importantly, each possible value of an input variable $x_i$ must be associated with a vector of unit norm~\cite{Stoudenmire2016}.
That is, each value that $x_i$ can assume must be represented by a vector in a higher-dimensional feature space.
Furthermore, to allow for a generative interpretation of the model, the feature vectors should be orthogonal~\cite{Stoudenmire2016}.
% Typically, this feature map is defined so that each of the vectors is orthogonal to the vectors associated with the other values.
This means that the vectors associated to the possible values of $x_i$ will form an orthonormal basis of the aforementioned feature space.
For a variable which can assume $n$ discrete values, this feature space will be $n$-dimensional.
The dimensionality of the space is referred to as the local dimension.

The unit norm and orthogonality conditions can be satisfied by defining a feature map $\phi^{(i)}(x_i)$, which maps each value onto a vector.
For example, if $x_i \in \{0, 1, 2\}$, a possible feature map is the one-hot encoding of each value: $(1, 0, 0)$ for 0, $(0, 1, 0)$ for 1, and $(0, 0, 1)$ for 2.
The feature map $\phi^{(i)}(x_i)$ allows us to rewrite the $A^{(i)}(x_i)$ in Eq.~\ref{eq:mps} as
\begin{equation}
    A^{(i)}_{\alpha_{i-1}\alpha_i}(x_i) = \sum_{\beta_i} T^{(i)}_{\alpha_{i-1}\beta_i\alpha_i} \phi^{(i)}_{\beta_i}(x_i),
\end{equation}
where $T^{(i)}_{\alpha_{i-1}\beta_i\alpha_i}$ is a tensor of rank 3.
Here, we have further decomposed $A^{(i)}$ into a contraction of tensor $T^{(i)}$ and the feature vector $\phi^{(i)}(x_i)$.
In graphical notation, the MPS (cf.\ Eq.~\ref{eq:mps}) becomes:
\begin{center}
    \begin{tikzpicture}[
    tensornode/.style={circle, draw=black!60, ultra thick, minimum size=8mm, inner sep=0},
    inputnode/.style={circle, draw=black!60, fill=gray!10, ultra thick, minimum size=7mm, inner sep=0},
    empty/.style={circle, ultra thick, minimum size=8mm, inner sep=0},
    ]
    %Nodes
    \node[rectangle] (text)   at (-1.25, 0.5) {$\Psi(x) =$};
    \node[tensornode](first)  at (0, 1)    {$T^{(1)}$};
    \node[tensornode](second) at (1.25, 1) {$T^{(2)}$};
    \node[tensornode](third)  at (2.5, 1)  {$T^{(3)}$};
    \node[empty]    (dotted)  at (3.5, 1)  {$\cdots$};
    \node[tensornode](nth)    at (4.5, 1)  {$T^{(n)}$};
    \node[inputnode] (finput) at (0, 0)    {$\phi^{(1)}$};
    \node[inputnode] (sinput) at (1.25, 0) {$\phi^{(2)}$};
    \node[inputnode] (tinput) at (2.5, 0)  {$\phi^{(3)}$};
    \node[inputnode] (ninput) at (4.5, 0)  {$\phi^{(n)}$};
    
    %Lines
    \draw[-, ultra thick] (first.east) -- (second.west);
    \draw[-, ultra thick] (second.east) -- (third.west);
    \draw[-, ultra thick] (third.east) -- (dotted.west);
    \draw[-, ultra thick] (dotted.east) -- (nth.west);
    \draw[-, ultra thick] (first.south) -- (finput.north);
    \draw[-, ultra thick] (second.south) -- (sinput.north);
    \draw[-, ultra thick] (third.south) -- (tinput.north);
    \draw[-, ultra thick] (nth.south) -- (ninput.north);
    \end{tikzpicture}
\end{center}

Given a data set, an MPS can be trained using a method based on the density matrix renormalization group (DMRG) algorithm \cite{Stoudenmire2016}.
This algorithm updates model parameters with respect to a given loss function by ``sweeping'' back and forth across the MPS.
In our case, we maximize the negative log-likelihood (NLL), i.e.\ we maximize the model evidence directly~\cite{Han2018}.
After training, the tensor network can be used to infer probability densities over unobserved variables by contracting the MPS with observed values.
For a more in-depth discussion on tensor networks, we refer to the Appendix.

\section{Environment}\label{ch:environment}
\begin{figure}[b]
    \centering
    \includegraphics[width=0.4\textwidth]{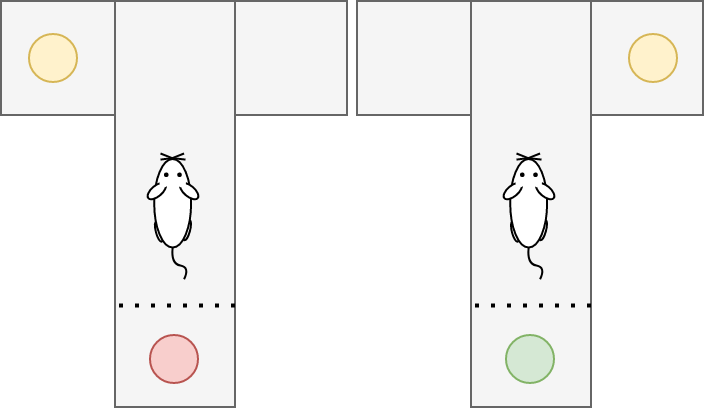}
    \caption{Possible contexts of the T-maze environment as presented by Friston et al.~\cite{Friston2016}. The agent starts off in the center. The reward (yellow) is located in either the left branch (left image) or right branch (right image). The cue reveals where the reward is: red indicates the reward is on the left, green indicates the reward is on the right.}
    \label{fig:mouse}
\end{figure}
The environment used to test the generative model is the classic T-maze as presented by Friston et al.~\cite{Friston2016}.
As the name suggests, the environment consists of a T-shaped maze and contains an artificial agent, e.g.\ a rat, and a reward, e.g.\ some cheese.
The maze is divided into four locations: the center, the right branch, the left branch, and the cue location.
The agent starts off in the center, while the reward is placed in either the left branch or the right branch, as depicted in Figure~\ref{fig:mouse}.
Crucially, the agent does not know the location of the reward initially.
Furthermore, once the agent chooses to go to either the left or the right branch, it is trapped and cannot leave.
For the agent, the initial state of the world is uncertain.
It does not know which type of world it is in: a world with the reward on the left, or a world with the reward on the right.
In other words, it does not know its context.
However, going to the cue reveals the location of the reward and enables the agent to observe the context.
This resolves all ambiguity and allows the agent to make the correct decision.

The implementation of the environment was provided by pymdp~\cite{Heins2022}.
In this package, the agent receives an observation with three modalities at every step: the location, the reward, and the context.
The location can take on four possible values: center, right, left, and cue, and indicates which location the agent is currently in.
The reward can take on three possible values: no reward, win, and loss.
The ``no reward'' observation indicates that the agent received no information about the reward, while the ``win'' and ``loss'' observations indicate that the agent either received the reward or failed to obtain the reward, respectively.
Logically, ``no reward'' can only be observed in the center and cue locations, while ``win'' and ``loss'' can only be observed in the left and right locations.
Finally, the context can take on two possible values: left and right.
Whenever the agent is in locations ``center'', ``left'' or ``right'', the context observation will be randomly selected from ``left'' or ``right'' uniformly.
Only when the agent is in the cue location, will the context observation yield the correct context.
Further, the possible actions that the agent can take include: center, right, left, and cue, corresponding to which location the agent wants to go to.

We modified the above implementation slightly to better reflect the environment brought forth by Friston et al.~\cite{Friston2016}.
In the original description, the agent is only allowed to perform two subsequent actions.
Therefore, the number of time steps was limited to two.
Furthermore, in the above implementation, the agent is able to leave the left and right branches of the T-maze.
Thus, we prevented the agent from leaving whenever it chose to go to the left or right branches.

\subsection{Modeling with tensor networks}\label{ch:tmazetn}
The tensor network was adapted in order to accommodate the environment and be able to receive sequences of actions and observations as input.
Firstly, the number of tensors was limited to the number of time steps.
Secondly, each tensor received an extra index so that the network may receive both actions and observations.
This led to the following network structure:
\begin{center}
    \begin{tikzpicture}[
    tensornode/.style={circle, draw=black!60, ultra thick, minimum size=8mm, inner sep=0},
    inputnode/.style={circle, draw=black!60, fill=gray!10, ultra thick, minimum size=7mm, inner sep=0},
    empty/.style={circle, ultra thick, minimum size=8mm, inner sep=0},
    ]
    %Nodes
    \node[rectangle] (text)   at (-1.25, 1) {$\Psi(x) =$};
    \node[tensornode](first)  at (0, 1)    {$T^{(1)}$};
    \node[tensornode](second) at (1.25, 1) {$T^{(2)}$};
    \node[tensornode](third)  at (2.5, 1)  {$T^{(3)}$};
    \node[inputnode] (finput) at (0, 0)    {$o_1$};
    \node[inputnode] (sinput) at (1.25, 0) {$o_2$};
    \node[inputnode] (tinput) at (2.5, 0)  {$o_3$};
    \node[inputnode] (sinput2) at (1.25, 2) {$a_1$};
    \node[inputnode] (tinput2) at (2.5, 2)  {$a_2$};
    
    %Lines
    \draw[-, ultra thick] (first.east) -- (second.west);
    \draw[-, ultra thick] (second.east) -- (third.west);
    \draw[-, ultra thick] (first.south) -- (finput.north);
    \draw[-, ultra thick] (second.south) -- (sinput.north);
    \draw[-, ultra thick] (third.south) -- (tinput.north);
    \draw[-, ultra thick] (second.north) -- (sinput2.south);
    \draw[-, ultra thick] (third.north) -- (tinput2.south);
    \end{tikzpicture}
\end{center}
where we used $a_i$ and $o_i$ to denote the feature vectors corresponding to action $a_i$ and observation $o_i$.
Note that the first tensor did not receive an action input, since we defined that action $a_i$ leads to observation $o_{i+1}$.

As mentioned in section~\ref{ch:tn}, the feature maps $\phi^{(i)}$ can generally be freely chosen as long as the resulting vectors are correctly normalized.
However, it is useful to select feature maps which can easily be inverted, such that feature space vectors can readily be interpreted.
% Moreover, for reasons based in quantum physics, it is good practice to make sure the resulting vectors are orthogonal.
In this case, since both observations and actions were discrete, one-hot encodings form a good option.
The feature vectors for actions were one-hot encodings of the possible actions.
The feature vectors for observations were one-hot encodings of the different combinations of observation modalities, i.e. $4 \times 3 \times 2 = 24$ one-hot vectors corresponding to different combinations of the three different modalities.
% Other feature maps are possible, however, the effect of the choice of feature map has not yet been extensively studied in the literature.

In principle, there is nothing stopping us from learning feature maps, as long as the maps are correctly normalized.
For practical purposes, the learning algorithm should make sure the feature maps are invertible.
Whether feature maps should be learned before training the model or can be learned on-the-fly is an open question.

At this point, it is important to mention that the feature map is not chosen with the intent of imposing structure on the feature space based on prior knowledge of the observations (or actions).
On the contrary, any feature map should assign a separate feature dimension to each possible observation, keeping the distance between that observation and all other observations within the feature space equal and maximal.
To this end, the feature map can be thought of as being a part of the likelihood.

\subsection{Planning with active inference}\label{ch:planning}
Once trained, the tensor network constructed in section \ref{ch:tmazetn} provides a generative model of the world.
In theory, this model can be used for planning with active inference.
At this point, it is important to remark that the network does not provide an accessible hidden state.
While the bonds between tensors can be regarded as internal states, they are not normalized and, therefore, not usable.
This poses a problem in the computation of the expected free energy, given by~\cite{Friston2016}
\begin{align}
    G(\pi) &= \sum_\tau G(\pi, \tau)\\
    G(\pi, \tau) &= \mathrm{E}_{Q(o_\tau | s_\tau, \pi)}[\log Q(s_\tau | \pi) - \log P(s_\tau, o_\tau | \Tilde{o}, \pi)]\\
    &\approx \underbrace{\mathrm{D_{KL}}(Q(o_\tau | \pi)\,||\,P(o_\tau))}_\text{expected cost} + \underbrace{\mathrm{E}_{Q(s_\tau | \pi)}[\mathrm{H}(Q(o_\tau | s_\tau, \pi))]}_\text{expected ambiguity},
    \label{eq:efe}
\end{align}
with hidden states $s_\tau$, observations $o_\tau$ and policy $\pi(\tau) = a_\tau$, where the $\sim$-notation denotes a sequence of variables $\Tilde{x} = (x_1, x_2, \ldots, x_{\tau-1})$ over time and $P(o_\tau)$ is the expected observation.
This computation requires access to the hidden state $s_\tau$ explicitly.

To remedy this, we suppose that the state $s_\tau$ contains all the information gathered across actions and observations that occur at times $t < \tau$.
Mathematically, we assume $Q(s_\tau | \pi) \approx Q(o_{<\tau} | \pi)$ and $Q(o_\tau | s_\tau, \pi) \approx Q(o_\tau | o_{<\tau}, \pi)$ with $o_{<\tau} = (o_1, \ldots, o_{(\tau-1)})$.
This way, we are able to approximate the expected ambiguity in Eq.~\ref{eq:efe}.
While these assumptions may give the impression that the calculation is computationally expensive, if the contraction with previous actions and observations has been performed once, it never has to be computed again, since the resulting tensor can be reused in subsequent calculations.
At this point, the resulting tensor contains all the information from previous actions and observations.
When planning, we must re-evaluate the likelihood (and thus the expected free energy) for every time step, while imposing that the previous time steps are fixed.
Indeed, we will perform sophisticated inference~\cite{Friston2021}.
Under this scheme, the expected free energy is given by
\begin{align}
    G(o_\tau, a_\tau) = \; &\underbrace{\mathrm{D_{KL}}(Q(o_{\tau+1} | a_{<\tau+1})\,||\,P(o_{\tau+1})) + \mathrm{E}_{Q(s_{\tau+1} | a_{<\tau+1})}[\mathrm{H}(P(o_{\tau+1} | s_{\tau+1}))]}_\text{expected free energy of next action}\nonumber\\
    &+ \underbrace{\mathrm{E}_{Q(a_{\tau+1} | o_{\tau+1})Q(o_{\tau+1}|a_{<\tau+1})}[G(o_{\tau+1}, a_{\tau+1})]}_\text{expected free energy of subsequent actions},\label{eq:si}\\
    Q(a_\tau | o_\tau) = \; &\sigma[-G(o_\tau, a_\tau)]
\end{align}
This defines a tree search over actions and observations in the future.

\section{Results and discussion}\label{ch:results}
In this section, we demonstrate how the model's beliefs shift over time.
Later, we show how a tensor network agent behaves when planning under sophisticated inference.

The data set was constructed by including one of every possible path through the maze, i.e. 202 sequences of actions and observations.
The model was trained over 500 epochs with a batch size of 10, where one epoch consisted of one right-to-left-to-right sweep per batch.
The learning rate was set to $10^{-4}$ and was further reduced by 10 \% whenever the loss increased too much (i.e.\ by more than $0.5$).
Additionally, bonds started with 8 dimensions.
The singular value cutoff point was set to 10 \% of the largest singular value.

\subsection{Belief shift}\label{ch:beliefs}
\begin{figure}[t!]
    \centering
    \includegraphics[height=1.5in,trim={0 0 0.55in 0},clip]{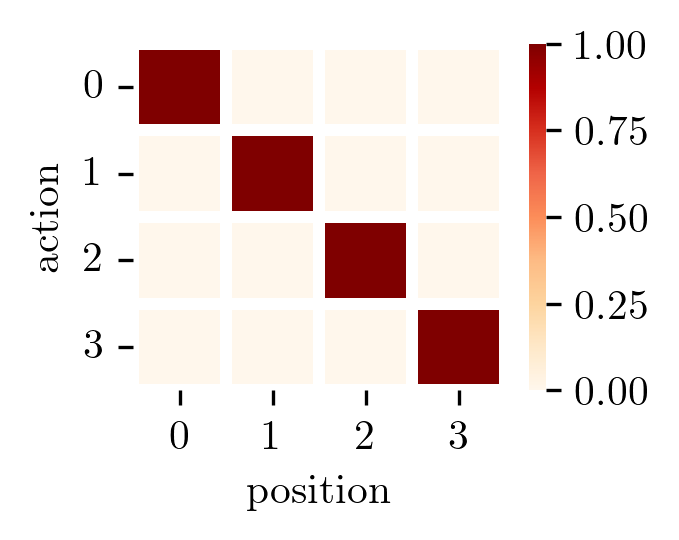}
    \includegraphics[height=1.5in,trim={0.4in 0 0.55in 0},clip]{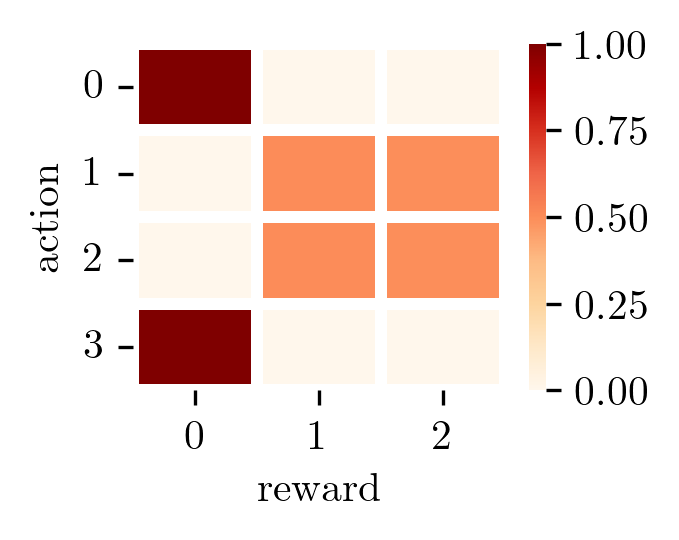}
    \includegraphics[height=1.5in,trim={0.4in 0 0 0},clip]{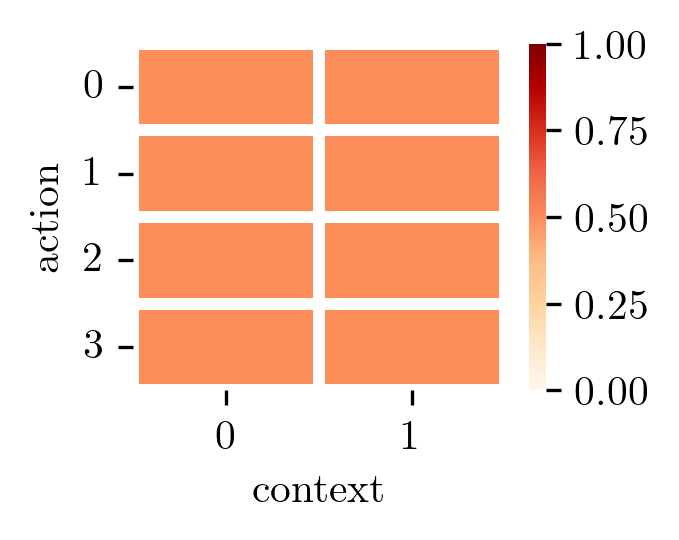}
    
    \includegraphics[height=1.5in,trim={0 0 0.55in 0},clip]{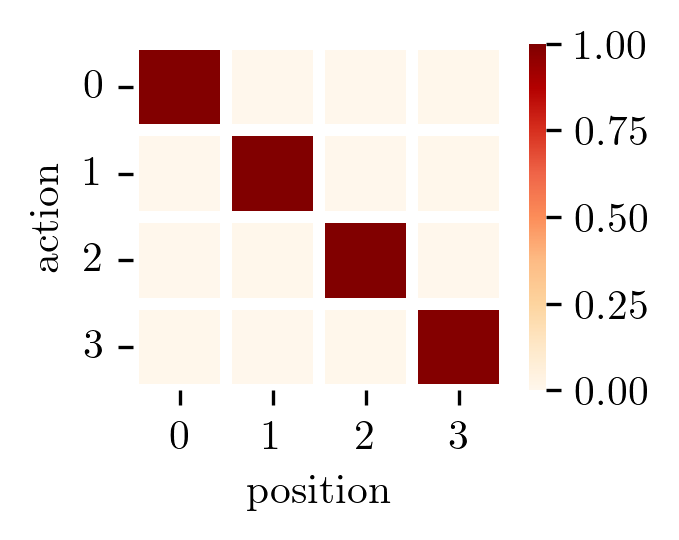}
    \includegraphics[height=1.5in,trim={0.4in 0 0.55in 0},clip]{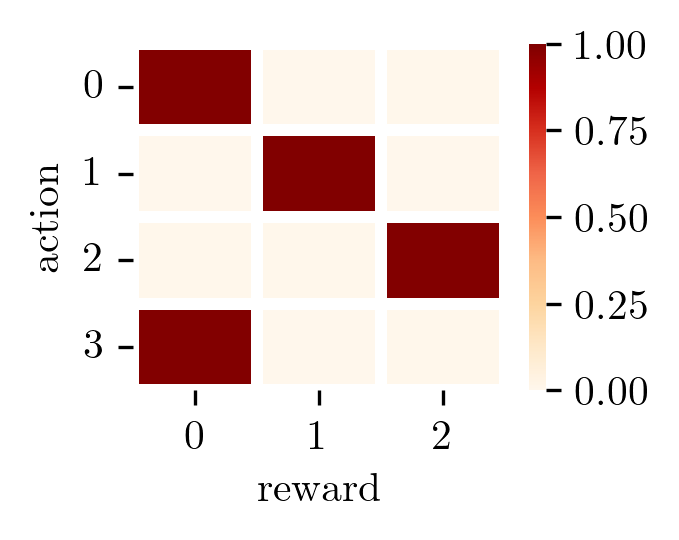}
    \includegraphics[height=1.5in,trim={0.4in 0 0 0},clip]{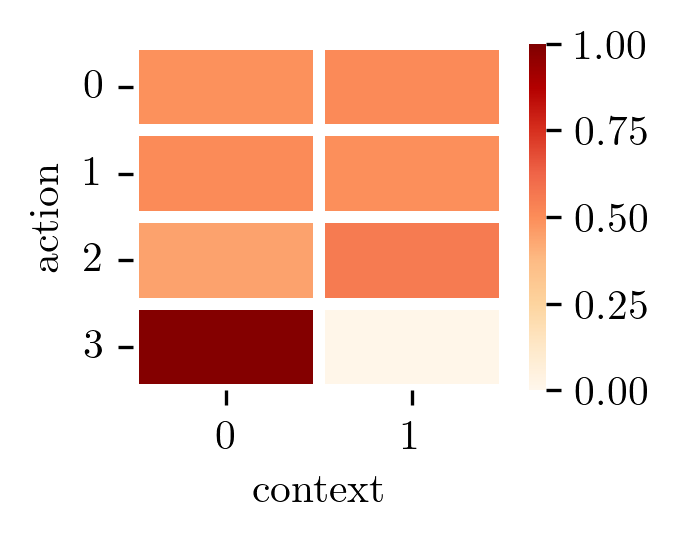}
    \caption{(top) Model beliefs over observation $o_2$ given action $a_1$ per modality. (bottom) Model beliefs for observation $o_3$ given action $a_2$ per modality, when the agent has observed the cue with context ``right''. Actions 0, 1, 2 and 3 correspond to center, right, left and cue, respectively. Positions 0, 1, 2 and 3 correspond to center, right, left and cue location, respectively. Rewards 0, 1 and 2 correspond to no reward, win and loss, respectively. Context 0 and 1 correspond to right and left, respectively.}
    \label{fig:beliefs}
\end{figure}
Since the initial observation $o_1$ is always center position, no reward and context right or left with 50 \% chance, we used the observation ``center, no reward and context right'' to obtain the beliefs in each case.
The results are analogous in the case of ``center, no reward and context left''.

Figure~\ref{fig:beliefs} (top) displays beliefs over $o_2$ given $a_1$.
From the results, it is clear that the agent does not know which reward it will receive, if it were to go to the left or right branch immediately.
In addition, it does not know which context it will observe, even if the agent were to go towards the cue.
Once the agent observes $o_2$, the beliefs shift.
Figure~\ref{fig:beliefs} (bottom) shows beliefs over $o_3$ given $a_2$ when the agent has seen the cue with context ``right``.
Since the agent has seen the cue, it is very certain about the reward it will receive if it goes to the left or the right branch.
If it stays in the cue location, it is also very certain that it will observe the same context again.

\subsection{Action selection}
With the outcome in section~\ref{ch:beliefs}, we were able to perform action selection based on the sophisticated inference scheme described in section~\ref{ch:planning}.
For this, we used the following preferred observation per modality:
\begin{equation}
    P(\text{position}) = \sigma(\begin{bmatrix}0 & 0 & 0 & 0\end{bmatrix}), \enspace
    P(\text{reward}) = \sigma(\begin{bmatrix}0 & 3 & -3\end{bmatrix}), \enspace
    P(\text{context}) = \sigma(\begin{bmatrix}0 & 0\end{bmatrix}).
    \label{eq:pref}
\end{equation}
\begin{figure}[t!]
    \centering
    \includegraphics[height=1.5in]{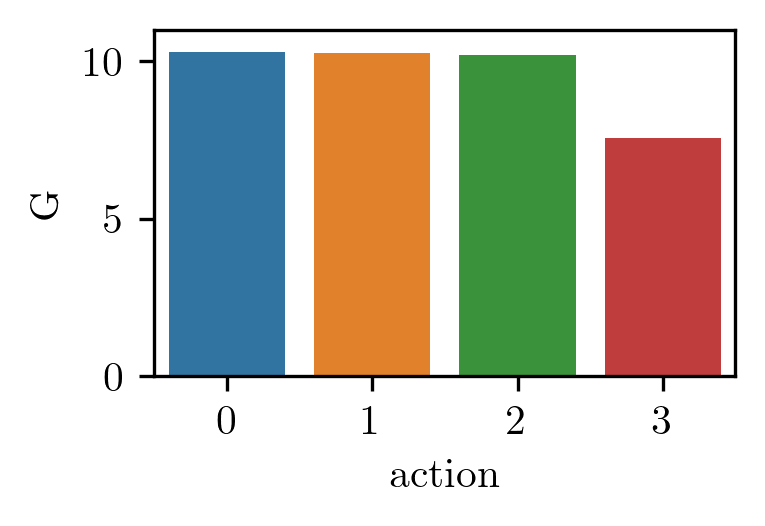}
    
    \includegraphics[height=1.5in]{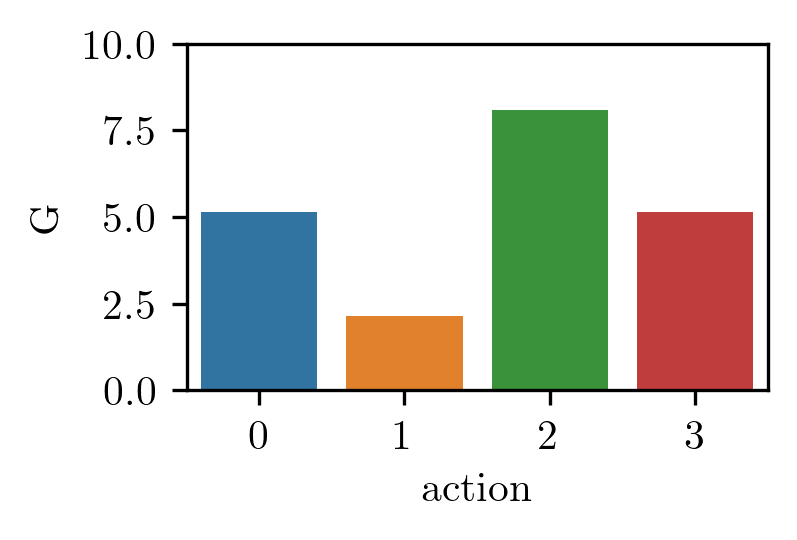}
    \includegraphics[height=1.5in]{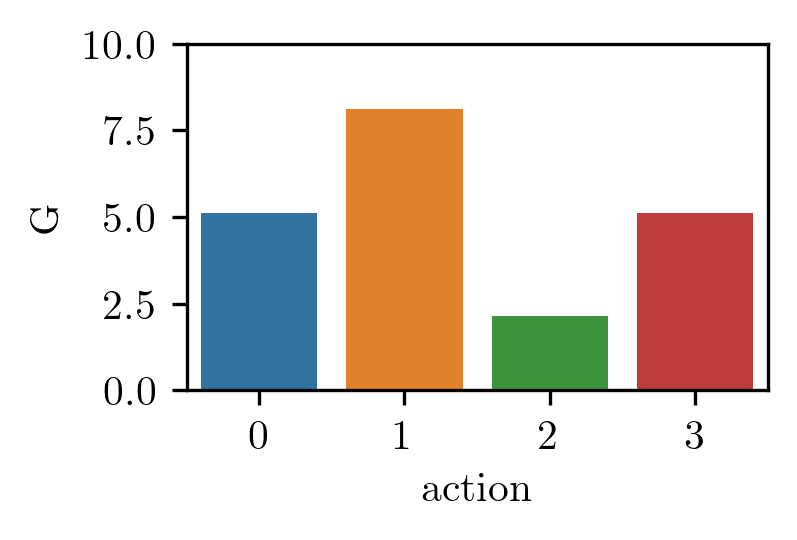}
    \caption{(top) Expected free energy for action $a_1$. (bottom) Expected free energy for action $a_2$ when observation $o_2$ was ``cue, no reward, context right'' and ``cue, no reward, context left'', respectively.}
    \label{fig:G}
\end{figure}
Figure~\ref{fig:G} (top) shows the expected free energy for action $a_1$.
Given that the expected free energy is lowest for the action that brings the agent to the cue, the agent will choose to go to the cue in the first action.
This is because, after observing the cue, the cue location provides a lower entropy on the context modality, as well as virtually 100\% certainty on where the reward is located.

Figure~\ref{fig:G} (bottom) shows the expected free energy for action $a_2$ when the agent has chosen to go to the cue location and has observed either ``cue, no reward, context right'' or ``cue, no reward, context left''.
In this case, the agent will choose to go to either the left or the right branch, depending on the context it observed, i.e.\ context right will lead to action right and vice-versa.

The net result is that the agent will first go to the cue in order to resolve ambiguity and, subsequently, go to the branch with the reward.

\section{Conclusion}\label{ch:conclusion}
We introduced a generative model based on tensor networks that is able to learn from sequential data.
In addition, we showed how one can obtain beliefs from such a generative model and how a (sophisticated) active inference agent can use these to compute the expected free energy.
Demonstration on the T-maze environment pointed out that such an agent is able to correctly select actions.

%future work
In the future, we plan to apply tensor networks to other environments, as well as make an in-depth comparison with neural networks, in order to better establish the benefits and drawbacks of the method.
Moreover, we will adapt the network to allow sequences of random lengths and look into incorporating observations with continuous variables, which may also allow us to undo the assumptions made in section~\ref{ch:planning}.
Both these changes will broaden the range of applicability of generative models based on tensor networks.

\section*{Acknowledgments}
This research received funding from the Flemish Government under the ``Onderzoeksprogramma Artifici\"ele Intelligentie (AI) Vlaanderen'' programme.
This work has received support from the European Union's Horizon 2020 program through Grant No.\ 863476 (ERC-CoG SEQUAM).

\bibliographystyle{splncs04}
\bibliography{mybibliography}

\clearpage
\begin{appendices}
\renewcommand{\thesection}{\appendixname~\arabic{section}}
\section{Notes on tensor networks}\label{app:tns}
The summation over a common index as in Eq.~\ref{eq:mps} is also called a contraction.
Performing the contraction between two tensors yields a new tensor with a rank equal to the sum of the ranks of the two contracted tensors minus two times the number of indices contracted over.
That is, contracting two tensors of rank 2 over a single index gives a new tensor of rank 2, which is simply matrix multiplication: $\sum_j A_{ij} B_{jk}$.
Similarly, contracting over the indices of a single tensor of rank 2 is simply the trace: $\sum_i A_{ii}$.
In this sense, contraction is a generalization of these operations.
Contracting indices in different ways gives rise to different structures.
Examples of other possible networks are: tree tensor networks (TTN) and projected entangled pair states (PEPS).

Tensor networks can more easily be understood using their graphical notation.
Each tensor is represented by a node, while contractions are represented by edges.
Free edges, i.e.\ edges which do not connect two tensors, correspond to free indices which have not been summed.
These can be used to represent sites in the network which are able to receive input or which can be used as input.

Some examples of tensors in graphical notation are:
\begin{itemize}
\item vector \quad
\begin{tikzpicture}[
baseline=-2,
tensornode/.style={circle, draw=black!60, ultra thick, minimum size=4mm, inner sep=0},
inputnode/.style={circle, draw=black!60, fill=gray!10, ultra thick, minimum size=7mm, inner sep=0},
empty/.style={circle, ultra thick, minimum size=1mm, inner sep=0},
line/.style={-, ultra thick}
]
%Nodes
\node[tensornode](first)   at (0, 0)     {};
\node[empty]     (leg1) at (0.5, 0)     {};

%Lines
\draw[line] (first.east) -- (leg1.west);
\end{tikzpicture},

\item matrix \quad
\begin{tikzpicture}[
baseline=-2,
tensornode/.style={circle, draw=black!60, ultra thick, minimum size=4mm, inner sep=0},
inputnode/.style={circle, draw=black!60, fill=gray!10, ultra thick, minimum size=7mm, inner sep=0},
empty/.style={circle, ultra thick, minimum size=1mm, inner sep=0},
line/.style={-, ultra thick}
]
%Nodes
\node[empty]     (leg1) at (0, 0)     {};
\node[tensornode](first)   at (0.5, 0)     {};
\node[empty]     (leg2) at (1, 0)     {};

%Lines
\draw[line] (leg1.east) -- (first.west);
\draw[line] (first.east) -- (leg2.west);
\end{tikzpicture},

\item rank-3 tensor \quad
\begin{tikzpicture}[
baseline=-2,
tensornode/.style={circle, draw=black!60, ultra thick, minimum size=4mm, inner sep=0},
inputnode/.style={circle, draw=black!60, fill=gray!10, ultra thick, minimum size=7mm, inner sep=0},
empty/.style={circle, ultra thick, minimum size=1mm, inner sep=0},
line/.style={-, ultra thick}
]
%Nodes
\node[empty]     (leg1) at (0, 0)     {};
\node[tensornode](first)   at (0.5, 0)     {};
\node[empty]     (leg2) at (1, 0)     {};
\node[empty]     (leg3) at (0.5, -0.5)     {};

%Lines
\draw[line] (leg1.east) -- (first.west);
\draw[line] (first.east) -- (leg2.west);
\draw[line] (first.south) -- (leg3.north);
\end{tikzpicture}.
\end{itemize}

Some examples of operations that can be represented by contractions are:

\begin{itemize}
\item dot product \quad
\begin{tikzpicture}[
baseline=-2,
tensornode/.style={circle, draw=black!60, ultra thick, minimum size=4mm, inner sep=0},
inputnode/.style={circle, draw=black!60, fill=gray!10, ultra thick, minimum size=7mm, inner sep=0},
empty/.style={circle, ultra thick, minimum size=1mm, inner sep=0},
line/.style={-, ultra thick}
]
%Nodes
\node[tensornode](first)   at (0.5, 0)     {};
\node[tensornode](second)   at (1.25, 0)     {};

%Lines
\draw[line] (first.east) -- (second.west);
\end{tikzpicture}\, ,

\item matrix multiplication \quad
\begin{tikzpicture}[
baseline=-2,
tensornode/.style={circle, draw=black!60, ultra thick, minimum size=4mm, inner sep=0},
inputnode/.style={circle, draw=black!60, fill=gray!10, ultra thick, minimum size=7mm, inner sep=0},
empty/.style={circle, ultra thick, minimum size=1mm, inner sep=0},
line/.style={-, ultra thick}
]
%Nodes
\node[empty]     (leg1) at (0, 0)     {};
\node[tensornode](first)   at (0.5, 0)     {};
\node[tensornode](second)   at (1.25, 0)     {};
\node[empty]     (leg3) at (1.75, 0)     {};

%Lines
\draw[line] (leg1.east) -- (first.west);
\draw[line] (first.east) -- (second.west);
\draw[line] (second.east) -- (leg3.west);
\end{tikzpicture},

\item trace \quad
\begin{tikzpicture}[
baseline=-2,
tensornode/.style={circle, draw=black!60, ultra thick, minimum size=4mm, inner sep=0},
inputnode/.style={circle, draw=black!60, fill=gray!10, ultra thick, minimum size=7mm, inner sep=0},
empty/.style={circle, ultra thick, minimum size=1mm, inner sep=0},
line/.style={-, ultra thick}
]
%Nodes
\node[tensornode](first)   at (0.5, 0)     {};

%Lines
\draw[line] (first.east) to[out=-45, in=45] (0.7, -0.35) to[out=-135, in=-45] (0.3, -0.35) to[out=135, in=-135] (first.west);
\end{tikzpicture}\, .
\end{itemize}

For a detailed account on tensor networks and their graphical notation, please refer to \cite{tns}.

\section{Training}\label{app:training}
The loss function must be chosen in such a way that the model captures the probability distribution of the data \cite{Han2018}.
A straightforward method for estimating the parameters of a probability distribution is maximum likelihood estimation.
In machine learning terms, this means we will optimize the parameters of the model with respect to the negative log-likelihood (NLL):
\begin{equation}
    \mathcal{L} = \frac{1}{|D|} \sum_{x \in D} \log P(x),
\end{equation}
where $D$ denotes the data set.
Through NLL minimization, the generative model becomes more similar to the probability distribution of the data.

\begin{figure}[t]
% \centering
\begin{tikzpicture}[
tensornode/.style={circle, draw=black!60, ultra thick, minimum size=4mm, inner sep=0},
inputnode/.style={circle, draw=black!60, fill=gray!10, ultra thick, minimum size=7mm, inner sep=0},
empty/.style={circle, ultra thick, minimum size=1mm, inner sep=0},
line/.style={-, ultra thick}
]
%Nodes
\node[rectangle](label)   at (-1, 0)     {a)};
\node[tensornode](first)   at (0, 0)     {};
\node[empty]     (dotted1) at (0.5, 0)     {$\cdots$};
\node[tensornode](im1th)   at (1, 0)  {};
\node[tensornode](ith)     at (1.75, 0)   {};
\node[tensornode](ip1th)   at (2.5, 0)  {};
\node[tensornode](ip2th)   at (3.25, 0)     {};
\node[empty]     (dotted2) at (3.75, 0)     {$\cdots$};
\node[tensornode](nth)     at (4.25, 0)     {};
\node[empty](e1)           at (0, -0.5)    {};
\node[empty](e2)           at (1, -0.5) {};
\node[empty](e3)           at (1.75, -0.5)  {};
\node[empty](e4)           at (2.5, -0.5) {};
\node[empty](e5)           at (3.25, -0.5)    {};
\node[empty](e6)           at (4.25, -0.5)    {};

%Lines
\draw[line] (first.east) -- (dotted1.west);
\draw[line] (dotted1.east) -- (im1th.west);
\draw[line] (im1th.east) -- (ith.west);
\draw[line] (ith.east) -- (ip1th.west);
\draw[line] (ip1th.east) -- (ip2th.west);
\draw[line] (ip2th.east) -- (dotted2.west);
\draw[line] (dotted2.east) -- (nth.west);
\draw[line] (first.south) -- (e1.north);
\draw[line] (im1th.south) -- (e2.north);
\draw[line] (ith.south) -- (e3.north);
\draw[line] (ip1th.south) -- (e4.north);
\draw[line] (ip2th.south) -- (e5.north);
\draw[line] (nth.south) -- (e6.north);
\end{tikzpicture}
\begin{tikzpicture}[
tensornode/.style={circle, draw=black!60, ultra thick, minimum size=4mm, inner sep=0},
inputnode/.style={circle, draw=black!60, fill=gray!10, ultra thick, minimum size=7mm, inner sep=0},
empty/.style={circle, ultra thick, minimum size=1mm, inner sep=0},
contr/.style={rectangle, draw=black!60, ultra thick, minimum width=12mm, minimum height=4mm, rounded corners},
line/.style={-, ultra thick}
]
%Nodes
\node[empty](start)        at (-1.1, 0) {};
\node[rectangle](text)     at (-0.75, 0)     {$=$};
\node[tensornode](first)   at (0, 0)     {};
\node[empty]     (dotted1) at (0.5, 0)     {$\cdots$};
\node[tensornode](im1th)   at (1, 0)  {};
\node[contr](b)            at (2.125, 0) {};
\node[tensornode](ip2th)   at (3.25, 0)     {};
\node[empty]     (dotted2) at (3.75, 0)     {$\cdots$};
\node[tensornode](nth)     at (4.25, 0)     {};
\node[empty](e1)           at (0, -0.5)    {};
\node[empty](e2)           at (1, -0.5) {};
\node[empty](e3)           at (1.75, -0.5)  {};
\node[empty](e4)           at (2.5, -0.5) {};
\node[empty](e5)           at (3.25, -0.5)    {};
\node[empty](e6)           at (4.25, -0.5)    {};

%Lines
\draw[line] (first.east) -- (dotted1.west);
\draw[line] (dotted1.east) -- (im1th.west);
\draw[line] (im1th.east) -- (b.west);
\draw[line] (b.east) -- (ip2th.west);
\draw[line] (ip2th.east) -- (dotted2.west);
\draw[line] (dotted2.east) -- (nth.west);
\draw[line] (first.south) -- (e1.north);
\draw[line] (im1th.south) -- (e2.north);
\draw[line] (b.south -| e3.north) -- (e3.north);
\draw[line] (b.south -| e4.north) -- (e4.north);
\draw[line] (ip2th.south) -- (e5.north);
\draw[line] (nth.south) -- (e6.north);
\end{tikzpicture}

\begin{tikzpicture}[
tensornode/.style={circle, draw=black!60, ultra thick, minimum size=4mm, inner sep=0},
inputnode/.style={circle, draw=black!60, fill=gray!10, ultra thick, minimum size=7mm, inner sep=0},
empty/.style={circle, ultra thick, minimum size=1mm, inner sep=0},
contr/.style={rectangle, draw=black!60, ultra thick, minimum width=12mm, minimum height=4mm, rounded corners},
line/.style={-, ultra thick}
]
%Nodes
\node[rectangle](label)   at (0, 0)     {b)};
\node[empty](im1th)    at (1, 0)       {};
\node[contr](b)        at (2.125, 0)   {$\Delta B$};
\node[empty](ip2th)    at (3.25, 0)    {};
\node[empty](e3)       at (1.75, -0.5) {};
\node[empty](e4)       at (2.5, -0.5)  {};
\node[rectangle](text) at (5, 0)       {$= \frac{Z^\prime}{Z} - \frac{2}{|D|} \sum_{x \in D} \frac{\Psi^\prime(x)}{\Psi(x)}$};

%Lines
\draw[line] (im1th.east) -- (b.west);
\draw[line] (b.east) -- (ip2th.west);
\draw[line] (b.south -| e3.north) -- (e3.north);
\draw[line] (b.south -| e4.north) -- (e4.north);
\end{tikzpicture}

\begin{tikzpicture}[
tensornode/.style={circle, draw=black!60, ultra thick, minimum size=4mm, inner sep=0},
inputnode/.style={circle, draw=black!60, fill=gray!10, ultra thick, minimum size=7mm, inner sep=0},
empty/.style={circle, ultra thick, minimum size=1mm, inner sep=0},
contr/.style={rectangle, draw=black!60, ultra thick, minimum width=12mm, minimum height=4mm, rounded corners},
line/.style={-, ultra thick},
]
%Nodes
\node[rectangle](label)   at (0, 0)     {c)};
\node[empty](im1th1)   at (1, 0)  {};
\node[contr](b1)       at (2.125, 0) {$B^\prime$};
\node[empty](ip2th1)   at (3.25, 0)     {};
\node[empty](e31)      at (1.75, -0.5)  {};
\node[empty](e41)      at (2.5, -0.5) {};
\node[rectangle](text1)    at (3.625, 0)     {$=$};
\node[empty](im1th2)   at (4, 0)  {};
\node[contr](b2)       at (5.125, 0) {$B$};
\node[empty](ip2th2)   at (6.25, 0)     {};
\node[empty](e32)      at (4.75, -0.5)  {};
\node[empty](e42)      at (5.5, -0.5) {};
\node[rectangle](text2)    at (6.75, 0)     {$+ \alpha$};
\node[empty](im1th3)   at (7, 0)  {};
\node[contr](b3)       at (8.125, 0) {$\Delta B$};
\node[empty](ip2th3)   at (9.25, 0)     {};
\node[empty](e33)      at (7.75, -0.5)  {};
\node[empty](e43)      at (8.5, -0.5) {};

%Lines
\draw[line] (im1th1.east) -- (b1.west);
\draw[line] (b1.east) -- (ip2th1.west);
\draw[line] (b1.south -| e31.north) -- (e31.north);
\draw[line] (b1.south -| e41.north) -- (e41.north);
\draw[line] (im1th2.east) -- (b2.west);
\draw[line] (b2.east) -- (ip2th2.west);
\draw[line] (b2.south -| e32.north) -- (e32.north);
\draw[line] (b2.south -| e42.north) -- (e42.north);
\draw[line] (im1th3.east) -- (b3.west);
\draw[line] (b3.east) -- (ip2th3.west);
\draw[line] (b3.south -| e33.north) -- (e33.north);
\draw[line] (b3.south -| e43.north) -- (e43.north);
\end{tikzpicture}

\begin{tikzpicture}[
tensornode/.style={circle, draw=black!60, ultra thick, minimum size=4mm, inner sep=0},
inputnode/.style={circle, draw=black!60, fill=gray!10, ultra thick, minimum size=7mm, inner sep=0},
empty/.style={circle, ultra thick, minimum size=1mm, inner sep=0},
contr/.style={rectangle, draw=black!60, ultra thick, minimum width=12mm, minimum height=4mm, rounded corners},
line/.style={-, ultra thick}
]
%Nodes
\node[rectangle](label)   at (0, 0)     {d)};
\node[empty](im1th)   at (1, 0)  {};
\node[contr](b)            at (2.125, 0) {$B^*$};
\node[empty](ip2th)   at (3.25, 0)     {};
\node[empty](e3)           at (1.75, -0.5)  {};
\node[empty](e4)           at (2.5, -0.5) {};
\node[empty](e1) at (4.375, 0) {};
\node[tensornode](u) at (5.125, 0) {$U$};
\node[tensornode](s) at (5.875, 0) {$S$};
\node[tensornode](v) at (6.625, 0) {$V$};
\node[empty](ue) at (5.125, -0.5) {};
\node[empty](ve) at (6.625, -0.5) {};
\node[empty](e2) at (7.25, 0) {};
\node[rectangle](eq) at (7.75, 0) {$=$};
\node[empty](e5) at (8.25, 0) {};
\node[tensornode](u2) at (9, 0) {};
\node[tensornode](v2) at (9.75, 0) {};
\node[empty](ue2) at (9, -0.5) {};
\node[empty](ve2) at (9.75, -0.5) {};
\node[empty](e6) at (10.5, 0) {};

%Lines
\draw[line] (im1th.east) -- (b.west);
\draw[line] (b.east) -- (ip2th.west);
\draw[line] (b.south -| e3.north) -- (e3.north);
\draw[line] (b.south -| e4.north) -- (e4.north);
\draw[line] (e1.east) -- (u.west);
\draw[line] (u.east) -- (s.west);
\draw[line] (s.east) -- (v.west);
\draw[line] (v.east) -- (e2.west);
\draw[line] (u.south) -- (ue.north);
\draw[line] (v.south) -- (ve.north);
\draw[line] (e5.east) -- (u2.west);
\draw[line] (u2.east) -- (v2.west);
\draw[line] (v2.east) -- (e6.west);
\draw[line] (u2.south) -- (ue2.north);
\draw[line] (v2.south) -- (ve2.north);
\draw[->, very thick, black!60, shorten >=2pt, shorten <=2pt] (ip2th.east) -- (e1.west) node [midway, above] {SVD};
\end{tikzpicture}
\caption{Training scheme for an MPS. a) Contraction of two adjacent tensors. b) Computing the update to the contracted tensor. c) Updating the contracted tensor. d) Decomposition of the contracted tensor using SVD.}
\label{fig:training}
\end{figure}
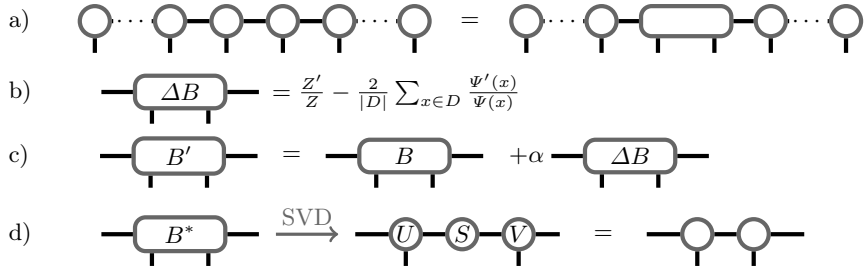

Training proceeds as depicted in Figure~\ref{fig:training}.
Firstly, tensor $T^{(i)}$ and $T^{(i+1)}$ are contracted to form the tensor $B^{(i, i+1)}$.
The update to $B^{(i, i+1)}$ is then computed using the loss function:
\begin{equation}
    \Delta B^{(i, i+1)} = \frac{\partial \mathcal{L}}{\partial B^{(i, i+1)}_{\alpha_{i-1} \beta_{i} \beta_{i+1} \alpha_{i+1}}} = \frac{Z^\prime}{Z} - \frac{2}{|D|} \sum_{x \in D} \frac{\Psi^\prime(x)}{\Psi(x)},
\end{equation}
where $Z^\prime = 2 \sum_{x \in D} \Psi^\prime(x) \Psi(x)$ and $\Psi^\prime$ is the derivative of $\Psi$ with respect to $B^{(i, i+1)}$.
Subsequently, the elements of $B^{(i, i+1)}$ are adjusted by adding $\Delta B^{(i, i+1)}$ multiplied by the learning rate.
Finally, the newly computed $B^{\prime(i, i+1)}$ is decomposed into two tensors again.
This decomposition is typically done through singular value decomposition (SVD), where the singular value matrix is then contracted with either the left or the right tensor, such that we are left with two tensors.

By starting this scheme at the leftmost tensor and iteratively moving one tensor to the right, the algorithm can update the entire MPS.
Indeed, it is possible to update one tensor $T^{(i)}_{\alpha_{i-1}\beta_i\alpha_{i}}$ at a time, however, the current method allows the dimensions of the indices $\alpha_{i}$ (graphically, the edges connecting $T^{(i)}$ nodes), the so-called bond dimensions, to vary during training.
This is made possible by truncated SVD, which truncates dimensions with singular values that fall beneath some manually specified threshold.
Truncating dimensions with small singular values can be interpreted as truncating less informative dimensions.
As a result, truncated SVD ensures that the model remains as small as possible, while containing the most information.
Moreover, the size of the model will vary depending on how much information it must learn.

\section{Computing probabilities with tensor networks}\label{app:probs}
One of the benefits of tensor networks is that we can easily obtain exact joint (and conditional) probability distributions without requiring parameterization.
After training the model, given that the network is correctly normalized ($Z=1$), the joint probability distribution is given by
\begin{center}
    \begin{tikzpicture}[
    tensornode/.style={circle, draw=black!60, ultra thick, minimum size=4mm, inner sep=0},
    inputnode/.style={circle, draw=black!60, fill=gray!10, ultra thick, minimum size=4mm, inner sep=0},
    empty/.style={circle, ultra thick, minimum size=8mm, inner sep=0},
    ]
    %Nodes
    \node[rectangle] (text)   at (-2.25, 0) {$P(x_1, x_2, x_3, \ldots, x_n) =$};
    \node[tensornode](first)  at (0, 0.9)    {};
    \node[tensornode](second) at (0.75, 0.9) {};
    \node[tensornode](third)  at (1.5, 0.9)  {};
    \node[empty]    (dotted)  at (2.25, 0.9)  {$\cdots$};
    \node[tensornode](nth)    at (3, 0.9)  {};
    \node[inputnode] (finput) at (0, 0.3)    {};
    \node[inputnode] (sinput) at (0.75, 0.3) {};
    \node[inputnode] (tinput) at (1.5, 0.3)  {};
    \node[inputnode] (ninput) at (3, 0.3)  {};
    
    \node[tensornode](first2)  at (0, -0.9)    {};
    \node[tensornode](second2) at (0.75, -0.9) {};
    \node[tensornode](third2)  at (1.5, -0.9)  {};
    \node[empty]    (dotted2)  at (2.25, -0.9)  {$\cdots$};
    \node[tensornode](nth2)    at (3, -0.9)  {};
    \node[inputnode] (finput2) at (0, -0.3)    {};
    \node[inputnode] (sinput2) at (0.75, -0.3) {};
    \node[inputnode] (tinput2) at (1.5, -0.3)  {};
    \node[inputnode] (ninput2) at (3, -0.3)  {};
    
    %Lines
    \draw[-, ultra thick] (first.east) -- (second.west);
    \draw[-, ultra thick] (second.east) -- (third.west);
    \draw[-, ultra thick] (third.east) -- (dotted.west);
    \draw[-, ultra thick] (dotted.east) -- (nth.west);
    \draw[-, ultra thick] (first.south) -- (finput.north);
    \draw[-, ultra thick] (second.south) -- (sinput.north);
    \draw[-, ultra thick] (third.south) -- (tinput.north);
    \draw[-, ultra thick] (nth.south) -- (ninput.north);
    
    \draw[-, ultra thick] (first2.east) -- (second2.west);
    \draw[-, ultra thick] (second2.east) -- (third2.west);
    \draw[-, ultra thick] (third2.east) -- (dotted2.west);
    \draw[-, ultra thick] (dotted2.east) -- (nth2.west);
    \draw[-, ultra thick] (first2.north) -- (finput2.south);
    \draw[-, ultra thick] (second2.north) -- (sinput2.south);
    \draw[-, ultra thick] (third2.north) -- (tinput2.south);
    \draw[-, ultra thick] (nth2.north) -- (ninput2.south);
    \end{tikzpicture}
\end{center}
where we have omitted tensor labels for simplicity.
Marginal distributions can be found by discarding the offending variables:
\begin{center}
    \begin{tikzpicture}[
    tensornode/.style={circle, draw=black!60, ultra thick, minimum size=4mm, inner sep=0},
    inputnode/.style={circle, draw=black!60, fill=gray!10, ultra thick, minimum size=4mm, inner sep=0},
    empty/.style={circle, ultra thick, minimum size=8mm, inner sep=0},
    ]
    %Nodes
    \node[rectangle] (text)   at (-3.8, 0) {$P(x_1) = \sum_{\{x\} \setminus \{x_1\}} P(x_1, x_2, x_3, \ldots, x_n) =$};
    \node[tensornode](first)  at (0, 0.9)    {};
    \node[tensornode](second) at (0.75, 0.9) {};
    \node[tensornode](third)  at (1.5, 0.9)  {};
    \node[empty]    (dotted)  at (2.25, 0.9)  {$\cdots$};
    \node[tensornode](nth)    at (3, 0.9)  {};
    \node[inputnode] (finput) at (0, 0.3)    {};
    
    \node[tensornode](first2)  at (0, -0.9)    {};
    \node[tensornode](second2) at (0.75, -0.9) {};
    \node[tensornode](third2)  at (1.5, -0.9)  {};
    \node[empty]    (dotted2)  at (2.25, -0.9)  {$\cdots$};
    \node[tensornode](nth2)    at (3, -0.9) {};
    \node[inputnode] (finput2) at (0, -0.3) {};
    
    %Lines
    \draw[-, ultra thick] (first.east) -- (second.west);
    \draw[-, ultra thick] (second.east) -- (third.west);
    \draw[-, ultra thick] (third.east) -- (dotted.west);
    \draw[-, ultra thick] (dotted.east) -- (nth.west);
    \draw[-, ultra thick] (first.south) -- (finput.north);
    
    \draw[-, ultra thick] (first2.east) -- (second2.west);
    \draw[-, ultra thick] (second2.east) -- (third2.west);
    \draw[-, ultra thick] (third2.east) -- (dotted2.west);
    \draw[-, ultra thick] (dotted2.east) -- (nth2.west);
    \draw[-, ultra thick] (first2.north) -- (finput2.south);
    \draw[-, ultra thick] (second2.north) -- (second.south);
    \draw[-, ultra thick] (third2.north) -- (third.south);
    \draw[-, ultra thick] (nth2.north) -- (nth.south);
    \end{tikzpicture}
\end{center}
where the sum over the variables $x_2, x_3, \ldots, x_n$ is equivalent to contracting the matching tensors.
Finally, conditional probability distributions can be found by combining the previous results:
\begin{equation}
    P(x_2, x_3, \ldots, x_n | x_1) = \frac{P(x_1, x_2, x_3, \ldots, x_n)}{P(x_1)}.
\end{equation}

\section{Physical intuition}\label{app:intuition}
In order to garner a feeling for the physics and mathematics used throughout this work, this section describes in (mostly) words what the physical meaning of the constituents of the tensor network is.

Let $x$ be an observable, i.e.\ a quantity that can be measured (or observed).
Examples of such physical quantities are position and momentum.
Furthermore, let $\{0, 1, 2\}$ be the set of values that $x$ can assume.

In quantum mechanics, the set of values that an observable can assume are eigenvalues.
This entails that there is a set of eigenvectors corresponding to these eigenvalues.
In turn, the eigenvectors form an eigenbasis of the state space.
In other words, every value that $x$ can assume has a corresponding vector and each of these vectors is a basis vector of the state space, meaning that each different value is represented in the state space by a different dimension.
For example, we may have the vectors $(1, 0, 0)$ for 0, $(0, 1, 0)$ for 1, and $(0, 0, 1)$ for 2.

Further, an MPS is typically used to represent a state within the state space.
This means that the MPS represents a (multi)vector.
This vector is not necessarily one of the eigenvectors mentioned earlier, but it can be virtually any vector within the state space.
When learning the parameters of an MPS, it is rotated and stretched in such a way that it represents the data.

If an MPS does not necessarily coincide with any of the eigenvectors, what value will it produce when a measurement is performed?
It will produce any of the eigenvalues with a certain probability.
These probabilities are given by the square of the inner product between the MPS and each eigenvector.
In our example, say the MPS was represented by the vector $(0.55, 0.55, 0.63)$, we would measure 0 with $30\%$ probability, 1 with $30\%$ probability and 2 with $40\%$ probability.

\end{appendices}

\end{document}